\def\year{2021}\relax
\documentclass[letterpaper]{article} 
\usepackage{aaai21}  
\usepackage{times}  
\usepackage{helvet} 
\usepackage{courier}  
\usepackage[hyphens]{url}  
\usepackage{graphicx} 
\usepackage{natbib}  
\usepackage{caption} 
\usepackage[super]{nth}
\usepackage{epsfig}
\usepackage{amsmath}
\usepackage{amssymb}
\usepackage{multirow}

\title{DeepFakesON-Phys: DeepFakes Detection based on Heart Rate Estimation}
\author{
    Javier Hernandez-Ortega, Ruben Tolosana, Julian Fierrez, Aythami Morales\\
}
\affiliations{
    Biometrics and Data Pattern Analytics Lab - BiDA Lab\\

    Universidad Autonoma de Madrid\\
    \{javier.hernandezo, ruben.tolosana, julian.fierrez, aythami.morales\}@uam.es

}

\begin{document}

\maketitle

\begin{abstract}
   This work introduces a novel DeepFake detection framework based on physiological measurement. In particular, we consider information related to the heart rate using remote photoplethysmography (rPPG). rPPG methods analyze video sequences looking for subtle color changes in the human skin, revealing the presence of human blood under the tissues. In this work we investigate to what extent rPPG is useful for the detection of DeepFake videos. The proposed fake detector named DeepFakesON-Phys uses a Convolutional Attention Network (CAN), which extracts spatial and temporal information from video frames, analyzing and combining both sources to better detect fake videos. DeepFakesON-Phys has been experimentally evaluated using the latest public databases in the field: Celeb-DF and DFDC. The results achieved, above 98\% AUC (Area Under the Curve) on both databases, outperform the state of the art and prove the success of fake detectors based on physiological measurement to detect the latest DeepFake videos.
\end{abstract}

\section{Introduction}


DeepFakes have become a great public concern recently~\cite{TED_news_concerns,BBC_doubleVideos}. The very popular term ``DeepFake" is usually referred to a deep learning based technique able to create fake videos by swapping the face of a person by the face of another person. This type of digital manipulation is also known in the literature as Identity Swap, and it is moving forward very fast~\cite{tolosana2020SurveyFakes}.

Currently, most face manipulations are based on popular machine learning techniques such as AutoEncoders (AE)~\cite{autoencoder_ICLR} and Generative Adversarial Networks (GAN)~\cite{goodfellow2014generative}, achieving in general very realistic visual results, specially in the latest generation of public DeepFakes~\cite{2020_Arxiv_DeepFakes_FaceRegions}, and the present trends~\cite{karras2019analyzing}. However, and despite the impressive visual results, are current face manipulations also considering the physiological aspects of the human being in the synthesis process?

Physiological measurement has provided very valuable information to many different tasks such as e-learning~\cite{JH_student_assessment}, health care~\cite{mcduff2015survey}, human-computer interaction~\cite{tan2010brain}, and security~\cite{handbook_Julian}, among many other tasks.

In physical face attacks, a.k.a. Presentation Attacks (PAs), real subjects are often impersonated using artifacts such as photographs, videos, and masks~\cite{handbook_Julian}. Face recognition systems are known to be vulnerable against these attacks unless proper detection methods are implemented~\cite{galbally14reviewAntispoofingFace,hernandez2019introduction}. Some of these detection methods are based on liveness detection by using information such as eye blinking or natural facial micro-expressions~\cite{bharadwaj2013computationally}.  Specifically for detecting 3D mask impersonation, which is one of the most challenging type of attacks, detecting pulse from face videos using remote photoplethysmography (rPPG) has shown to be an effective countermeasure~\cite{hernandez2018time}. When applying this technique to a video sequence with a fake face, the estimated heart rate signal is significantly different to the heart rate extracted from a real face~\cite{erdogmus2014spoofing}. 

\begin{figure*}[t]
\begin{center}
\includegraphics[width=0.9\linewidth]{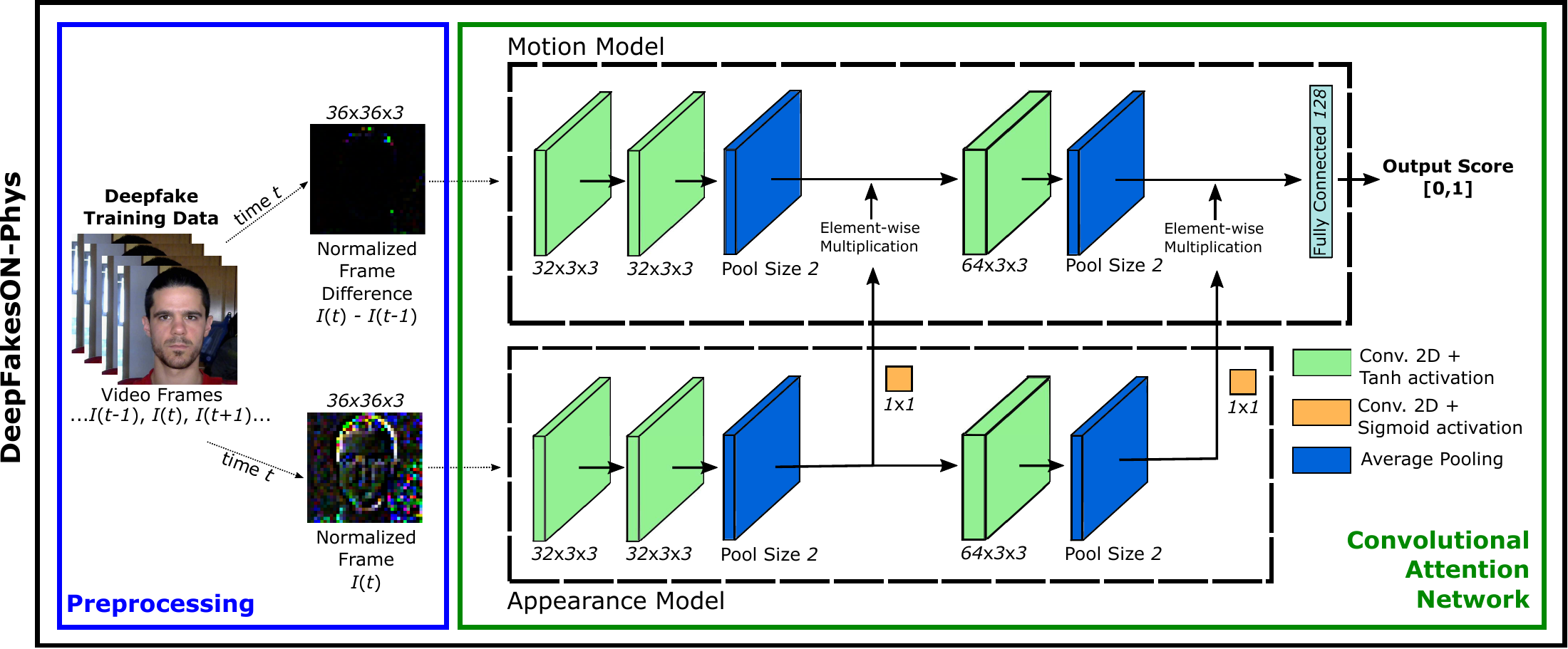}
\end{center}
\caption{\textbf{DeepFakesON-Phys architecture}. It comprises two stages: \textit{i)} a preprocessing step to normalize the video frames, and \textit{ii)} a Convolutional Attention Network composed of Motion and Appearance Models to better detect fake videos.}
\label{proposed_figure}
\end{figure*}

Seeing the good results achieved by rPPG techniques when dealing with physical 3D face mask attacks, and since DeepFakes are digital manipulations somehow similar to them, in this work we hypothesize that fake detectors based on physiological measurement can also be used against DeepFakes after adapting them properly. DeepFake generation methods have historically tried to mimic the visual appearance of genuine faces. However, to the best of our knowledge, they do not emulate the physiology of human beings, e.g., heart rate, blood oxygenation, or breath rate, so estimating that type of signals from the video could be a powerful tool for the detection of DeepFakes. 


The \textbf{novelty of this work consists in using rPPG features previously learned for the task of heart rate estimation and adapting them for the detection of DeepFakes by means of a knowledge-transfer process, thus obtaining a novel fake detector based on physiological measurement named DeepFakesON-Phys.} In particular, the information related to the heart rate is considered to decide whether a video is real or fake. Our physiological detector intends to be a robust solution to the weaknesses of most state-of-the-art DeepFake detectors based on the visual features existing in fake videos~\cite{matern2019exploiting,2019_Agarwal} and also on the artifacts/fingerprints inserted during the synthesis process~\cite{2019_Arxiv_GANRemoval_Tolosana}, which are highly dependent on a specific fake manipulation technique. 

DeepFakesON-Phys is based on DeepPhys \cite{chen2018deepphys}, a deep learning model trained for heart rate estimation from face videos based on rPPG. DeepPhys showed high accuracy even when dealing with challenging conditions such as heterogeneous illumination or low resolution, outperforming classic hand-crafted approaches. We used the architecture of DeepPhys, but making changes to make it suitable for DeepFake detection. We initialized the weights of the layers of DeepFakesON-Phys with the ones from DeepPhys (meant for heart rate estimation based on rPPG) and we adapted them to the new task using fine-tuning. This process allowed us to train our detector without the need of a high number of samples (compared to training it from scratch). Fine-tuning also helped us to obtain a model that detects DeepFakes by looking to rPPG related features from the images in the face videos.

\begin{table*}[t!]
\centering
\vspace{2mm}
\caption{\textbf{Comparison of different state-of-the-art fake detectors.} Results in \textit{italics} indicate the generalization capacity of the detectors against unseen databases. FF++ = FaceForensics++, AUC = Area Under the Curve, Acc. = Accuracy, EER = Equal Error Rate.}
\label{table:relatedWorks_faceSwapping}
\resizebox{0.90\textwidth}{!}{
\begin{tabular}{ccccc}
\textbf{Study}                                                                                 & \textbf{Method}                       & \textbf{Classifiers}                & \textbf{Best Performance}                                                                       & \textbf{Databases}                                                                                                   \\ \hline \hline

\multirow{4}{*}{\begin{tabular}[c]{@{}c@{}}\\  \cite{matern2019exploiting}\end{tabular}}   & \multirow{4}{*}{Visual Features}       & \multirow{4}{*}{\begin{tabular}[c]{@{}c@{}}\\Logistic Regression\\MLP\end{tabular}}        & AUC = 85.1\%                                                                                    & Own   \\ \cline{4-5}
                                                                                              &                                         &                                     & \textit{AUC = 78.0\%}                                                                           & \textit{FF++ / DFD}                                                                                                 
                                                                                              \\ \cline{4-5}
                                                                                              &                                         &                                     & \textit{AUC = 66.2\%}                                                                           & \textit{DFDC Preview}                                                                                                  \\ \cline{4-5}
                                                                                              &                                         &                                     & \textit{AUC = 55.1\%}                                                                           & \textit{Celeb-DF}                                                                                                    \\ \hline \hline

\multirow{4}{*}{\begin{tabular}[c]{@{}c@{}}\\ \cite{Li2019CVPR,li2019celebdf}\end{tabular}}       & \multirow{4}{*}{Face Warping Features} & \multirow{4}{*}{CNN}                & AUC = 97.7\% & UADFV                                                                                                                                  \\ \cline{4-5}
                                                                                              &                                         &                                     & \textit{AUC = 93.0\%}                                                                           & \textit{FF++ / DFD}                                                                                                 
                                                                                              \\ \cline{4-5}
                                                                                              &                                         &                                     & \textit{AUC = 75.5\%}                                                                           & \textit{DFDC Preview}                                                                                                  \\ \cline{4-5}
                                                                                              &                                         &                                     & \textit{AUC = 64.6\%}                                                                           & \textit{Celeb-DF}                                                                                                    \\ \hline \hline

\multirow{4}{*}{\begin{tabular}[c]{@{}c@{}} \cite{rossler2019faceforensics++}\end{tabular}} & \multirow{4}{*}{\begin{tabular}[c]{@{}c@{}} Mesoscopic Features \\ Steganalysis Features \\ Deep Learning Features \end{tabular}} & \multirow{4}{*}{CNN}                
& \begin{tabular}[c]{@{}c@{}}Acc. $\simeq$ 94.0\%\\ Acc. $\simeq$ 98.0\%\\ Acc. $\simeq$ 100.0\%\end{tabular}
& \begin{tabular}[c]{@{}c@{}}FF++ (DeepFake, LQ)\\ FF++ (DeepFake, HQ)\\ FF++ (DeepFake, RAW)\end{tabular} \\ \cline{4-5}
&                                         &                            
& \begin{tabular}[c]{@{}c@{}}Acc. $\simeq$ 93.0\%\\ Acc. $\simeq$ 97.0\%\\ Acc. $\simeq$ 99.0\%\end{tabular}  
& \begin{tabular}[c]{@{}c@{}}FF++ (FaceSwap, LQ)\\FF++ (FaceSwap, HQ)\\ FF++ (FaceSwap, RAW)\end{tabular}    \\ \hline \hline

                                                                                              \multirow{4}{*}{\begin{tabular}[c]{@{}c@{}}\\  \cite{nguyen2019use}\end{tabular}}  & \multirow{4}{*}{Deep Learning Features} & \multirow{4}{*}{Capsule Networks}        & \textit{AUC = 61.3\%}                                                                           & \textit{UADFV}                                                                                                                                    \\ \cline{4-5}
                                                                                              &                                         &                                     & \textit{AUC = 96.6\%}                                                                           & \textit{FF++ / DFD}                                                                                                  \\ \cline{4-5}
                                                                                              &                                         &                                     & \textit{AUC = 53.3\%}                                                                                    & \textit{DFDC Preview}
                                                                                              \\ \cline{4-5}
                                                                                              &                                         &                                     & \textit{AUC = 57.5\%}                                                                                    & \textit{Celeb-DF}                                                                                                                                                                                                 \\ \hline
\begin{tabular}[c]{@{}c@{}} \cite{Jain2019facialManipulation}\end{tabular}               & Deep Learning Features                  & CNN + Attention Mechanism                 & \begin{tabular}[c]{@{}c@{}}AUC = 99.4\%\\ EER = 3.1\%\end{tabular}                     & DFFD                                                                                                        \\ \hline
\begin{tabular}[c]{@{}c@{}} \cite{dolhansky2019deepfake}\end{tabular}               & Deep Learning Features & CNN                                 & \begin{tabular}[c]{@{}c@{}}Precision = 93.0\%\\ Recall = 8.4\%\end{tabular}              & DFDC Preview                                                                                        \\ \hline \hline

\begin{tabular}[c]{@{}c@{}} \cite{2019_Sabir}\end{tabular}     & {\begin{tabular}[c]{@{}c@{}}Image + Temporal Features\end{tabular}}     & CNN + RNN          & {\begin{tabular}[c]{@{}c@{}}AUC = 96.9\% \\AUC = 96.3\%\end{tabular}} & {\begin{tabular}[c]{@{}c@{}}FF++ (DeepFake, LQ)\\FF++ (FaceSwap, LQ)\end{tabular}} \\ \hline \hline

\multirow{4}{*}{\begin{tabular}[c]{@{}c@{}} \cite{2020_Arxiv_DeepFakes_FaceRegions}\end{tabular}}  & \multirow{4}{*}{Facial Regions Features} & \multirow{4}{*}{CNN}        & AUC = 100.0\%                                                                           & UADFV                                                                                                                                    \\ \cline{4-5}
                                                                                              &                                         &                                     & AUC = 99.5\%                                                                           & FF++ (FaceSwap, HQ)                                                                                                                                                                                                 \\ \cline{4-5}
                                                                                              &                                         &                                     & AUC = 91.1\%                                                                                    & DFDC Preview                                                                                                
                                                                                              \\ \cline{4-5}
                                                                                              &                                         &                                     & AUC = 83.6\%                                                                                    & Celeb-DF
                                                                                                \\ \hline \hline

\begin{tabular}[c]{@{}c@{}} \cite{conotter2014physiologically}\end{tabular}     & Physiological Features     & -          & Acc. = 100\% & Own                                                                          \\ \hline                                                                                                
                                                                                                \begin{tabular}[c]{@{}c@{}} \cite{li2018ictu}\end{tabular}     & Physiological Features     & LRCN          & AUC = 99.0\% & UADFV                                                                          \\ \hline

\begin{tabular}[c]{@{}c@{}} \cite{2019_Agarwal}\end{tabular}     & Physiological Features     & SVM          & AUC = 96.3\% & Own (FaceSwap, HQ)                                                                          \\ \hline

\begin{tabular}[c]{@{}c@{}} \cite{ciftci2019fakecatcher}\end{tabular}     & Physiological Features     & SVM/CNN          & {\begin{tabular}[c]{@{}c@{}}Acc. = 94.9\% \\Acc. = 91.5\%\end{tabular}} & {\begin{tabular}[c]{@{}c@{}}FF++ (DeepFakes)\\Celeb-DF\end{tabular}} \\ \hline 

\begin{tabular}[c]{@{}c@{}} \cite{eyeBlinking_2020}\end{tabular}     & Physiological Features     & Distance          & Acc. = 87.5\% & Own                                                                           \\ \hline

\multirow{3}{*}{\begin{tabular}[c]{@{}c@{}}\\  \cite{DeepRythm_deepfakes}\end{tabular}}  & \multirow{3}{*}{Physiological Features} & \multirow{3}{*}{CNN + Attention Mechanism}        & Acc. = 100.0\%                                                                           & FF++ (FaceSwap) \\ \cline{4-5}
                                                                                              &                                         &                                     & Acc. = 100.0\%                                                                          & FF++ (DeepFake) \\ \cline{4-5}
                                                                                              &                                         &                                     & \textit{Acc. = 64.1\%}                                                                                    & \textit{DFDC Preview}

                                                                                                \\ \hline \hline

\multirow{2}{*}{\begin{tabular}[c]{@{}c@{}} \textbf{DeepFakesON-Phys} \textbf{[Ours]}\end{tabular}}  & \multirow{2}{*}{\textbf{Physiological Features}} & \multirow{2}{*}{\textbf{CAN}}        & \textbf{AUC = 99.9\%}                                                                                    & \textbf{Celeb-DF v2} \\ \cline{4-5}
                                                                                              &                                         &                                     & \textbf{AUC = 98.2\%}                                                                                    & \textbf{DFDC Preview}

                                                                                                \\ \hline \hline

\end{tabular}

}
\end{table*}

In this context, the main contributions of our work are:
\begin{itemize}

\item \textbf{An in-depth literature review of DeepFake detection} approaches with special emphasis to physiological techniques, including the key aspects of the detection systems, the databases used, and the main results achieved.

\item \textbf{An approach based on physiological measurement to detect DeepFake videos: DeepFakesON-Phys\footnote{https://github.com/BiDAlab/DeepFakesON-Phys}.} Fig.~\ref{proposed_figure} graphically summarizes the proposed fake detection approach based on the original architecture DeepPhys \cite{chen2018deepphys}, a Convolutional Attention Network (CAN) composed of two parallel Convolutional Neural Networks (CNN) able to extract spatial and temporal information from video frames. This architecture is adapted for the detection of DeepFake videos by means of a knowledge-transfer process.

\item \textbf{A thorough experimental assessment of the proposed DeepFakesON-Phys}, considering the latest public databases of the \nth{2} DeepFake generation such as Celeb-DF v2 and DFDC Preview. DeepFakesON-Phys achieves high-accuracy results, outperforming the state of the art. In addition, the results achieved prove that current face manipulation techniques do not pay attention to the heart-rate-related physiological information of the human being when synthesizing fake videos.




\end{itemize}

The remainder of the paper is organized as follows. \textbf{Related Works} summarizes previous studies focused on the detection of DeepFakes. \textbf{Proposed Method: DeepFakesON-Phys} describes the proposed DeepFakesON-Phys fake detection approach. \textbf{Databases} summarizes all databases considered in the experimental framework of this study. \textbf{Experiments} describes the experimental protocol and the results achieved in comparison with the state of the art. Finally, \textbf{Conclusions} draws the final conclusions and points out future research lines.

\section{Related Works}\label{relatedWorks}
Different approaches have been proposed in the literature to detect DeepFake videos. Table~\ref{table:relatedWorks_faceSwapping} shows a comparison of the most relevant approaches in the area, paying special attention to the fake detectors based on physiological measurement. For each study we include information related to the method, classifiers, best performance, and databases for research. It is important to remark that in some
cases, different evaluation metrics are considered, e.g., Area Under the Curve (AUC) and Equal Error Rate (EER), which complicate the comparison among studies. Finally, the results highlighted in \textit{italics} indicate the generalization ability of the detectors against unseen databases, i.e., those databases were not considered
for training. Most of these results are extracted from~\cite{li2019celebdf}.

The first studies in the area focused on the visual artifacts existed in the \nth{1} generation of fake videos. The authors of~\cite{matern2019exploiting} proposed fake detectors based on simple visual artifacts such as eye colour, missing reflections, and missing details in the teeth areas, achieving a final 85.1\% AUC.

Approaches based on the detection of the face warping artifacts have also been studied in the literature. For example,~\cite{Li2019CVPR,li2019celebdf} proposed detection systems based on CNN in order to detect the presence of such artifacts from the face and the surrounding areas, being one of the most robust detection approaches against unseen face manipulations. 

Undoubtedly, fake detectors based on pure deep learning features are the most popular ones: feeding the networks with as many real/fake videos as possible and letting the networks to automatically extract the discriminative features. In general, these fake detectors have achieved very good results using popular network architectures such as Xception~\cite{rossler2019faceforensics++,dolhansky2019deepfake}, novel ones such as Capsule Networks~\cite{nguyen2019use}, and novel training techniques based on attention mechanisms~\cite{Jain2019facialManipulation}.

Fake detectors based on the image and temporal discrepancies across frames have also been proposed in the literature.~\cite{2019_Sabir} proposed a Recurrent Convolutional Network similar to~\cite{2018_David_AVSS}, trained end-to-end instead of using a pre-trained model. Their proposed detection approach was tested using FaceForensics++ database~\cite{rossler2019faceforensics++}, achieving AUC results above 96\%. 

Although most approaches are based on the detection of fake videos using the whole face, in~\cite{2020_Arxiv_DeepFakes_FaceRegions} the authors evaluated the discriminative power of each facial region using state-of-the-art network architectures, achieving interesting results on DeepFake databases of the \nth{1} and \nth{2} generations.

Finally, we pay special attention to the fake detectors based on physiological information. The eye blinking rate was studied in~\cite{li2018ictu,eyeBlinking_2020}.~\cite{li2018ictu} proposed Long-Term Recurrent Convolutional Networks (LRCN) to capture the temporal dependencies existed in human eye blinking. Their method was evaluated on the UADFV database, achieving a final 99.0\% AUC. More recently, ~\cite{eyeBlinking_2020} proposed a different approach named DeepVision. They fused the Fast-HyperFace~\cite{ranjan2017hyperface} and EAR~\cite{soukupova2016eye} algorithms to track the blinking, achieving an accuracy of 87.5\% over an in-house database.

Fake detectors based on the analysis of the way we speak were studied in~\cite{2019_Agarwal}, focusing on the distinct facial expressions and movements. These features were considered in combination with Support Vector Machines (SVM), achieving a 96.3\% AUC over their own database.

Finally, fake detection methods based on the heart rate have been also studied in the literature. One of the first studies in this regard was~\cite{conotter2014physiologically} where the authors preliminary evaluated the potential of blood flow changes in the face to distinguish between computer generated and real videos. Their proposed approach was evaluated using 12 videos (6 real and fake videos each), concluding that it is possible to use this metric to detect computer generated videos. 

Changes in the blood flow have also been studied in~\cite{ciftci2019fakecatcher,DeepRythm_deepfakes} using DeepFake videos. In~\cite{ciftci2019fakecatcher}, the authors considered rPPG techniques to extract robust biological features. Classifiers based on SVM and CNN were analyzed, achieving final accuracies of 94.9\% and 91.5\% for the DeepFakes videos of FaceForensics++ and Celeb-DF, respectively.

Recently, in~\cite{DeepRythm_deepfakes} a more sophisticated fake detector named DeepRhythm was presented. This approach was also based on features extracted using rPPG techniques. DeepRhythm was enhanced through two modules: \textit{i)} motion-magnified spatial-temporal representation, and \textit{ii)} dual-spatial-temporal attention. These modules were incorporated in order to provide a better adaptation to dynamically changing faces and various fake types. In general, good results with accuracies of 100\% were achieved on FaceForensics++ database. However, this method suffers from a demanding preprocessing stage, needing a precise detection of $81$ facial landmarks and the use of a color magnification algorithm prior to fake detection. Also, poor results were achieved on databases of the \nth{2} generation such as the DFDC Preview (Acc. = 64.1\%).



In the present work, in addition to the proposal of a different DeepFake detection architecture, we enhance previous approaches, e.g.~\cite{DeepRythm_deepfakes}, by keeping the preprocessing stage as light and robust as possible, only composed of a face detector and frame normalization. To provide an overall picture, we include in Table~\ref{table:relatedWorks_faceSwapping} the results achieved with our proposed DeepFakesON-Phys in comparison with key related works, which shows that we outperform the state of the art on Celeb-DF v2 and DFDC Preview databases.

\section{Proposed Method: DeepFakesON-Phys}\label{proposedApproach}
Fig.~\ref{proposed_figure} graphically summarizes the architecture of DeepFakesON-Phys, the proposed fake detector based on heart rate estimation. We hypothesize that rPPG methods should obtain significantly different results when trying to estimate the subjacent heart rate from a video containing a real face, compared with a fake face. Since the changes in color and illumination due to oxygen concentration are subtle and invisible to the human eye, we think that most of the existing DeepFake manipulation methods do not consider the physiological aspects of the human being yet.



The initial architecture of DeepFakesON-Phys is based on the DeepPhys model described in \cite{chen2018deepphys}, whose objective was to estimate the human heart rate using facial video sequences. The model is based on deep learning and was designed to extract spatio-temporal information from videos mimicking the behavior of traditional handcrafted rPPG techniques. Features are extracted through the color changes in users’ faces that are caused by the variation of oxygen concentration in the blood. Signal processing methods are also used for isolating the color changes caused by blood from other changes that may be caused by factors such as external illumination, noise, etc. 

As can be seen in Fig.~\ref{proposed_figure}, after the first preprocessing stage, the Convolutional Attention Network (CAN) is composed of two different CNN branches:

\begin{itemize}
\item \textbf{Motion Model}: it is designed to detect changes between consecutive frames, i.e., performing a short-time analysis of the video for detecting fakes. To accomplish this task, the input at a time \textit{t} consists of a frame computed as the normalized difference of the current frame $I(t)$ and the previous one $I(t-1)$.

\item \textbf{Appearance Model}: it focuses on the analysis of the static information on each video frame. It has the target of providing the Motion Model with information about which points of the current frame may contain the most relevant information for detecting DeepFakes, i.e., a batch of attention masks that are shared at different layers of the CNN. The input of this branch at time $t$ is the raw frame of the video $I(t)$, normalized to zero mean and unitary standard deviation. 
\end{itemize} 

The attention masks coming from the Appearance Model are shared with the Motion Model at two different points of the CAN. Finally, the output layer of the Motion Model is also the final output of the entire CAN. 

In the original architecture \cite{chen2018deepphys}, the output stage consisted of a regression layer for estimating the time derivative of the subject's heart rate. In our case, as we do not aim to estimate the pulse of the subject, but the presence of a fake face, we change the final regression layer to a classification layer, using a sigmoid activation function for obtaining a final score in the [$0$,$1$] range for each instant $t$ of the video, related to the probability of the face being real.

Since the original DeepPhys model from \cite{chen2018deepphys} is not publicly available, instead of training a new CAN from scratch, we decided to initialize DeepFakesON-Phys with the weights from the model pretrained for heart rate estimation presented in~\cite{hernandez2020comparative}, which is also an adaptation of DeepPhys but trained using the COHFACE database~\cite{heusch2017reproducible}. This model also showed to have high accuracy in the heart rate estimation task using real face videos, so our idea is to take benefit of that acquired knowledge to better train DeepFakesON-Phys through a proper fine-tuning process.

Once we initialized DeepFakesON-Phys with the mentioned weights, we freeze the weights of all the layers of the original CAN model apart from the new classification layer and the last fully-connected layer, and we retrain the model. Due to this fine-tuning process we take benefit of the weights learned for heart rate estimation, just adapting them for the DeepFake detection task. This way, we make sure that the weights of the convolutional layers remain looking for information relative to heart rate and the last layers learn how to use that information for detecting the existence of DeepFakes.

\section{Databases}\label{databases}
Two different public databases are considered in the experimental framework of this study. In particular, Celeb-DF v2 and DFDC Preview, the two most challenging DeepFake databases up to date. Their videos exhibit a large range of variations in aspects such as face sizes (in pixels), lighting conditions (i.e., day, night, etc.), backgrounds, different acquisition scenarios (i.e., indoors and outdoors), distances from the person to the camera, and pose variations, among others.
These databases present enough images (fake and genuine) to fine-tune the original weights meant for heart rate estimation, obtaining new weights also based in rPPG features but adapted for DeepFake detection. Table~\ref{table:databases_faceSwap} summarizes the main characteristics of the databases.

\subsection{Celeb-DF v2}\label{celeb_database}
The aim of the Celeb-DF v2 database~\cite{li2019celebdf} was to generate fake videos of better visual quality compared with the previous UADFV database. This database consists of $590$ real videos extracted from Youtube, corresponding to celebrities with a diverse distribution in terms of gender, age, and ethnic group. Regarding fake videos, a total of $5$,$639$ videos were created swapping faces using DeepFake technology. The final videos are in MPEG4.0 format.

\begin{figure*}[t]
\centering
\includegraphics[width=0.85\linewidth]{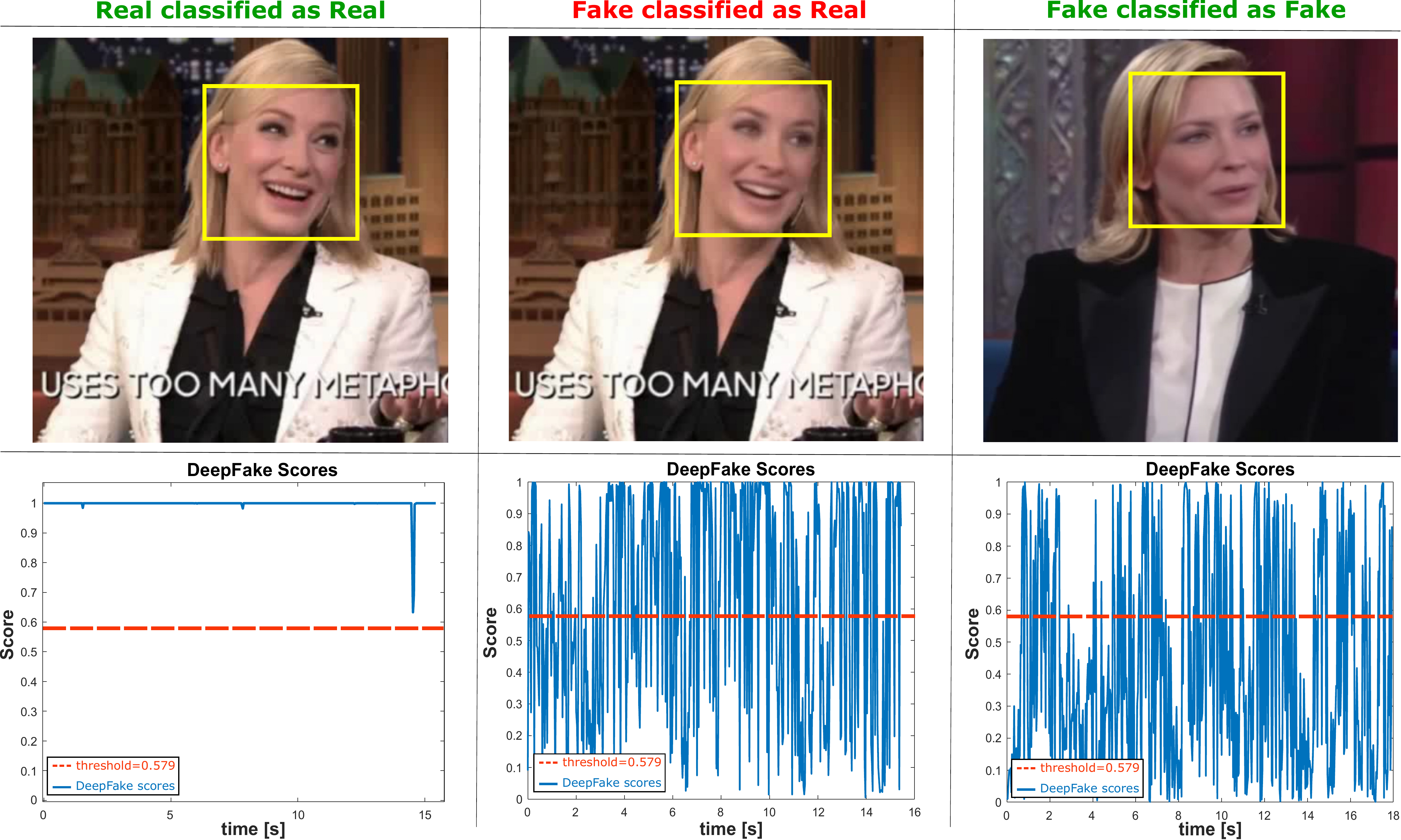}
\caption{\textbf{Examples of successful and failed DeepFake detections}. Top: sample frames of evaluated videos. Bottom: score distribution for each sample video. For the fake video misclassified as containing a real face, the DeepFake detection scores present a higher mean compared to the case of the fake video correctly classified as a fake.}
\label{fail_cases}
\end{figure*}

\begin{table}[t]
\centering
\vspace{2mm}
\caption{\textbf{Identity swap publicly available databases} of the \nth{2} generation considered in our experimental framework.}
\label{table:databases_faceSwap}
\scalebox{0.9}{
\begin{tabular}{ccc}
\multicolumn{3}{c}{\textbf{\nth{2} Generation}} \\ \hline
\textbf{Database}                                                              & \textbf{Real Videos} & \textbf{Fake Videos}                                                          \\ \hline \hline                                                            
\begin{tabular}[c]{@{}c@{}}Celeb-DF v2 \\ \cite{li2019celebdf}\end{tabular}          & 590 (Youtube)        & 5,639 (DeepFake) \\ \hline  
\begin{tabular}[c]{@{}c@{}}DFDC Preview \\ \cite{dolhansky2019deepfake}\end{tabular}     & 1,131 (Actors)        & 4,119 (Unknown)                                                          \\ \hline
\end{tabular}
}

\end{table}

\subsection{DFDC Preview}\label{DFDC_database}
The DFDC database~\cite{dolhansky2019deepfake} is one of the latest public databases, released by Facebook in collaboration with other companies and academic institutions such as Microsoft, Amazon, and the MIT. In the present study we consider the DFDC Preview dataset consisting of $1$,$131$ real videos from $66$ paid actors, ensuring realistic variability in gender, skin tone, and age. It is important to remark that no publicly available data or data from social media sites were used to create this dataset, unlike other popular databases. Regarding fake videos, a total of $4$,$119$ videos were created using two different unknown approaches for fakes generation. Fake videos were generated by swapping subjects with similar appearances, i.e., similar facial attributes such as skin tone, facial hair, glasses, etc. After a given pairwise model was trained on two identities, the identities were swapped onto the other’s videos.

\section{Experiments}\label{experimentalFramework}

\subsection{Experimental Protocol}\label{experimentalProtocol}
Celeb-DF v2 and DFDC Preview databases have been divided into non-overlapping datasets, development and evaluation. It is important to remark that each dataset comprises videos from different identities (both real and fake), unlike some previous studies. This aspect is very important in order to perform a fair evaluation and predict the generalization ability of the fake detection systems against unseen identities. Also, it is important to remark that the evaluation is carried out at frame level as in most previous studies~\cite{tolosana2020SurveyFakes}, not video level, using the popular AUC and accuracy metrics.

For the Celeb-DF v2 database, we consider real/fake videos of 40 and 19 different identities for the development and evaluation datasets respectively, whereas for the DFDC Preview database, we follow the same experimental protocol proposed in~\cite{dolhansky2019deepfake} as the authors already considered this concern.

\subsection{Fake Detection Results: DeepFakesON-Phys}\label{experimentalResults}

%

This section evaluates the ability of DeepFakesON-Phys to detect the most challenging DeepFake videos of the \nth{2} generation. Table~\ref{evaluation_results} shows the fake detection performance results achieved in terms of AUC and accuracy over the final evaluation datasets of Celeb-DF v2 and DFDC Preview. It is important to highlight that a separate fake detector is trained for each database. 

In general, very good results are achieved in both DeepFake databases. For the Celeb-DF v2 database, DeepFakesON-Phys achieves an accuracy of $98.7\%$ and an AUC of $99.9\%$. Regarding the DFDC Preview database, the results achieved are $94.4\%$ accuracy and $98.2\%$ AUC, similar ones to the obtained for the Celeb-DF database.


%

Observing the results, it seems clear that the fake detectors have learnt to distinguish the spatio-temporal differences between the real/fake faces of Celeb-DF v2 and DFDC Preview databases. Since all the convolutional layers of the proposed fake detector are frozen (the network was originally initialized with the weights from the model trained to predict the heart rate~\cite{hernandez2020comparative}), and we only train the last fully-connected layers, we can conclude that the proposed detection approach based on physiological measurement is successfully using pulse-related features for distinguishing between real and fake faces. These results prove that current face manipulation techniques do not pay attention to the heart-rate-related physiological information of the human being when synthesizing fake videos.

\begin{table}[t]
\centering
\vspace{2mm}
\caption{\textbf{Fake detection performance} results in terms of AUC and Accuracy over the final evaluation datasets.}
\label{evaluation_results}
\scalebox{0.95}{
\begin{tabular}{ccc}
\textbf{Database}     & \textbf{AUC Results (\%)} & \textbf{Acc. Results (\%)} \\ \hline \hline
Celeb-DF v2  & 99.9             & 98.7              \\
DFDC Preview & 98.2             & 94.4             
\end{tabular}
}

\end{table}

\begin{table*}[t]
\centering
\vspace{2mm}
\caption{\textbf{Comparison of different state-of-the-art fake detectors with our proposed DeepFakesON-Phys.} The best results achieved for each database are remarked in \textbf{bold}. Results in \textit{italics} indicate that the evaluated database (Celeb-DF or DFDC) was not used for training.}
\label{table:comparison_state_art}
\resizebox{1\textwidth}{!}{
\begin{tabular}{ccccc}
\multirow{2}{*}{\textbf{Study}} & \multirow{2}{*}{\textbf{Method}}         & \multirow{2}{*}{\textbf{Classifiers}}       & \multicolumn{2}{c}{\textbf{AUC Results (\%)}}                                                                                                                  \\ \cline{4-5}
                      &                                 &                                    & \textbf{Celeb-DF}~\cite{li2019celebdf}                                                       & \textbf{DFDC}~\cite{dolhansky2019deepfake}                                                           \\  \hline \hline \\
~\cite{yang2019exposing}            & Head Pose Features              & SVM                                & \textit{54.6}                                                             & \textit{55.9}                                                             \\
~\cite{li2019celebdf}              & Face Warping Features           & CNN                                & \textit{64.6}                                                             & \textit{75.5}                                                             \\
~\cite{afchar2018mesonet}          & Mesoscopic Features             & CNN                                & \textit{54.8}                                                             & \textit{75.3}                                                             \\
~\cite{Jain2019facialManipulation}            & Deep Learning Features          & CNN + Attention Mechanism          & \textit{71.2}                                                             & -                                                                         \\
~\cite{2020_Arxiv_DeepFakes_FaceRegions}        & Deep Learning Features          & CNN                                & 83.6                                                                      & 91.1                                                                      \\
~\cite{DeepRythm_deepfakes}              & Physiological Features          & CNN + Attention Mechanism          & \textit{-}                                                                & \textit{Acc. = 64.1}                                                      \\
~\cite{ciftci2019fakecatcher}          & Physiological Features          & SVM/CNN                            & Acc. = 91.5                                                               & -                                                                         \\ \\ \hline \\
\begin{tabular}[c]{@{}c@{}}\textbf{DeepFakesON-Phys} \textbf{[Ours]}\end{tabular} & \textbf{Physiological Features} & \textbf{CNN + Attention Mechanism} & \textbf{\begin{tabular}[c]{@{}c@{}}AUC = 99.9\\ Acc. = 98.7\end{tabular}} & \textbf{\begin{tabular}[c]{@{}c@{}}AUC = 98.2\\ Acc. = 94.4\end{tabular}}
\end{tabular}

}

\end{table*}

Fig.~\ref{fail_cases} shows some examples of successful and failed detections when evaluating the proposed approach with real/fake faces of Celeb-DF v2. In particular, all the failures correspond to fake faces generated from a particular video, misclassifying them as real faces. Fig.~\ref{fail_cases} shows a frame from the original real video (top-left), one from a misclassified fake video generated using that scenario (top-middle), and another from a fake video correctly classified as fake and generated using the same real and fake identities but from other source videos (top-right). The detection threshold is the same for all the testing databases and videos, and it has been selected to maximize the accuracy in the evaluation.

Looking at the score distributions along time of the three examples (Fig.~\ref{fail_cases}, bottom), it can be seen that for the real face video (left) the scores are $1$ for most of the time and always over the detection threshold. However, for the fake videos considered (middle and right), the score changes constantly, making the score of some fake frames to cross the detection threshold and consequently misclassifying them as real. Nevertheless, it is important to remark that these mistakes only happen if we analyze the results at frame level (traditional approach followed in the literature~\cite{tolosana2020SurveyFakes}). In case we consider an evaluation at video level, DeepFakesON-Phys would be able to detect fake videos by integrating the temporal information available in short-time segments, e.g., in a similar way as described in~\cite{hernandez2018time} for continuous face anti-spoofing.

We believe that the failures produced in this particular case are propitiated by the interferences of external illumination. rPPG methods that use handcrafted features are usually fragile against external artificial illumination in the frequency and power ranges of normal human heart rate, making difficult to distinguish those illumination changes from the color changes caused by blood perfusion. Anyway, the proposed physiological approach presented in this work is more robust to this kind of illumination perturbations than hand-crafted methods, thanks to the fact that the training process is data-driven, making possible to identify those interferences by using their presence in the training data.

\subsection{Comparison with the State of the Art}
Finally, we compared in Table~\ref{table:comparison_state_art} the results achieved in the present work with other state-of-the-art DeepFake detection approaches: head pose variations~\cite{yang2019exposing}, face warping artifacts~\cite{li2019celebdf}, mesoscopic features~\cite{afchar2018mesonet}, pure deep learning features~\cite{Jain2019facialManipulation,2020_Arxiv_DeepFakes_FaceRegions}, and physiological features~\cite{DeepRythm_deepfakes,ciftci2019fakecatcher}. The best results achieved for each database are remarked in \textbf{bold}. Results in \textit{italics} indicate that the evaluated database was not used for training. Some of these results are extracted from~\cite{li2019celebdf}.

Note that the comparison in Table~\ref{table:comparison_state_art} is not always under the same datasets and protocols, therefore it must be interpreted with care. Despite of that, it is patent that the proposed DeepFakesON-Phys has achieved state-of-the-art results in both Celeb-DF and DFDC Preview databases. In particular, it has further outperformed popular fake detectors based on pure deep learning approaches such as Xception and Capsule Networks~\cite{2020_Arxiv_DeepFakes_FaceRegions} and also other recent physiological approaches based on SVM/CNN~\cite{ciftci2019fakecatcher}.

\section{Conclusions}\label{conclusions}

This work has evaluated the potential of physiological measurement to detect DeepFake videos. In particular, we have proposed a novel DeepFake detector named DeepFakesON-Phys based on a Convolutional Attention Network (CAN) originally trained for heart rate estimation using remote photoplethysmography (rPPG). The proposed CAN approach consists of two parallel CNN networks that extract and share temporal and spatial information from video frames. 

DeepFakesON-Phys has been evaluated using Celeb-DF v2 and DFDC Preview databases, two of the latest and most challenging DeepFake video databases. Regarding the experimental protocol, each database was divided into development and evaluation datasets, considering different identities in each dataset in order to perform a fair evaluation of the technology.

The soundness and competitiveness of DeepFakesON-Phys has been proven by the very good results achieved, AUC values of 99.9\% and 98.2\% for the Celeb-DF and DFDC databases, respectively. These results have outperformed other state-of-the-art fake detectors based on face warping and pure deep learning features, among others. Finally, the experimental results of this study reveal that current face manipulation techniques do not pay attention to the heart-rate-related or blood-related physiological information.


Immediate work may consist in replicating the state of the art DeepFake works and training them with the same databases than the ones used to train DeepFakesON-Phys in order to make a fair comparison of accuracy, and showing the actual performance of our method. Another future work will be oriented to the analysis of the robustness of the proposed fake detection approach against face manipulations unseen during the training process~\cite{tolosana2020SurveyFakes}, temporal integration of frame data~\cite{hernandez2018time}, and the application of the proposed physiological approach to other face manipulation techniques such as face morphing~\cite{raja2020morphing}. 

\section{Acknowledgments}
This work has been supported by projects: IDEA-FAST (IMI2-2018-15-two-stage-853981), PRIMA (ITN-2019-860315), TRESPASS-ETN (ITN-2019-860813), BIBECA (RTI2018-101248-B-I00 MINECO/FEDER), and edBB (Universidad Autonoma de Madrid, UAM). J. H.-O. is supported by a PhD fellowship from UAM. R. T. is supported by a Postdoctoral fellowship from CAM/FSE.

\begin{quote}
\begin{small}
\bibliography{egbib}
\end{small}
\end{quote}

\end{document}